\documentclass[conference]{IEEEtran}
\IEEEoverridecommandlockouts

\usepackage{amsmath,amssymb,amsfonts}
\usepackage{subcaption}
\usepackage{algorithm}
\usepackage{algorithmic}
\usepackage{acronym}
\usepackage{graphicx}
\usepackage{textcomp}
\usepackage{float}
\usepackage{subcaption}
\usepackage{xcolor}
\def\BibTeX{{\rm B\kern-.05em{\sc i\kern-.025em b}\kern-.08em
    T\kern-.1667em\lower.7ex\hbox{E}\kern-.125emX}}

\usepackage[backend=biber, sorting=none, style=numeric-comp, maxnames=5]{biblatex}
\DeclareSourcemap{
  \maps[datatype=bibtex]{
    \map[overwrite=true]{
      \step[fieldset=url, null]
      \step[fieldset=eprint, null]
    }
  }
}

  \graphicspath{{img/} {figures/} {figures_SM/} {./}}
\DeclareGraphicsExtensions{.pdf,.png,.svg,.jpg,.mps}
    \acrodef{AOI}[AOI]{Areas of Interest}
\acrodef{ATC}[ATC]{Air Traffic Control} 
\acrodef{ADC}[ADC]{Analog-to-Digital Converter}
\acrodef{ADEXP}[AdExp-IF]{Adaptive Exponential Integrate-and-Fire}
\acrodef{ADM}[ADM]{Asynchronous Delta Modulator}
\acrodef{AE}[AE]{Address-Event}
\acrodef{AER}[AER]{Address-Event Representation}
\acrodef{AEX}[AEX]{AER EXtension board}
\acrodef{AFE}[AFE]{Analog Front-End}
\acrodef{AFM}[AFM]{Atomic Force Microscope}
\acrodef{AGC}[AGC]{Automatic Gain Control}
\acrodef{AI}[AI]{Artificial Intelligence}
\acrodef{AMDA}[AMDA]{AER Motherboard with D/A converters}
\acrodef{AMPA}[AMPA]{$\alpha$-Amino-3-hydroxy-5-methyl-4-isoxazolepropionic Acid}
\acrodef{ANN}[ANN]{Artificial Neural Network}
\acrodef{API}[API]{Application Programming Interface}
\acrodef{APMOM}[APMOM]{Alternate Polarity Metal On Metal}
\acrodef{ARM}[ARM]{Advanced RISC Machine}
\acrodef{ASIC}[ASIC]{Application Specific Integrated Circuit}
\acrodef{BCM}[BMC]{Bienenstock-Cooper-Munro}
\acrodef{BD}[BD]{Bundled Data}
\acrodef{BEOL}[BEOL]{Back-end of Line}
\acrodef{BG}[BG]{Bias Generator}
\acrodef{BMI}[BMI]{Brain-Machince Interface}
\acrodef{BTB}[BTB]{Band-to-Band tunnelling}
\acrodef{bpm}[bpm]{Beats per Minute}
\acrodef{CA}[CA]{Cortical Automaton}
\acrodef{CAD}[CAD]{Computer Aided Design}
\acrodef{CAM}[CAM]{Content Addressable Memory}
\acrodef{CAVIAR}[CAVIAR]{Convolution AER Vision Architecture for Real-Time}
\acrodef{CCN}[CCN]{Cooperative and Competitive Network}
\acrodef{CDR}[CDR]{Clock-Data Recovery}
\acrodef{CFC}[CFC]{Current to Frequency Converter}
\acrodef{CHP}[CHP]{Communicating Hardware Processes}
\acrodef{CMIM}[CMIM]{Metal-Insulator-Metal Capacitor}
\acrodef{CML}[CML]{Current Mode Logic}
\acrodef{CMOL}[CMOL]{Hybrid CMOS nanoelectronic circuits}
\acrodef{CMOS}[CMOS]{Complementary Metal-Oxide-Semiconductor}
\acrodef{CNN}[CNN]{Convolutional Neural Network}
\acrodef{CNS}[CNS]{central Nervous System}
\acrodef{COTS}[COTS]{Commercial Off-The-Shelf}
\acrodef{CPG}[CPG]{Central Pattern Generator}
\acrodef{CPLD}[CPLD]{Complex Programmable Logic Device}
\acrodef{CPU}[CPU]{Central Processing Unit}
\acrodef{CSM}[CSM]{Cortical State Machine}
\acrodef{CSP}[CSP]{Constraint Satisfaction Problem}
\acrodef{CTXCTL}[CTXCTL]{CortexControl}
\acrodef{CV}[CV]{Coefficient of Variation}
\acrodef{DAC}[DAC]{Digital to Analog Converter}
\acrodef{DAS}[DAS]{Dynamic Auditory Sensor}
\acrodef{DAVIS}[DAVIS]{Dynamic and Active Pixel Vision Sensor}
\acrodef{DBN}[DBN]{Deep Belief Network}
\acrodef{DBS}[DBS]{Deep Brain Stimulation}
\acrodef{DFA}[DFA]{Deterministic Finite Automaton}
\acrodef{DIBL}[DIBL]{Drain-Induced Barrier-Lowering}
\acrodef{DI}[DI]{Delay Insensitive}
\acrodef{divmod3}[DIVMOD3]{Divisibility of a number by three}
\acrodef{DMA}[DMA]{Direct Memory Access}
\acrodef{DNF}[DNF]{Dynamic Neural Field}
\acrodef{DNN}[DNN]{Deep Neural Network}
\acrodef{DOF}[DOF]{Degrees of Freedom}
\acrodef{DPE}[DPE]{Dynamic Parameter Estimation}
\acrodef{DPI}[DPI]{Differential Pair Integrator}
\acrodef{DRAM}[DRAM]{Dynamic Random Access Memory}
\acrodef{DR}[DR]{Dual Rail}
\acrodef{DRRZ}[DR-RZ]{Dual-Rail Return-to-Zero}
\acrodef{DSP}[DSP]{Digital Signal Processor}
\acrodef{DVS}[DVS]{Dynamic Vision Sensor}
\acrodef{DYNAP-SE}[DYNAP-SE]{Dynamic Neuromorphic Asynchronous Processor}
\acrodef{EBL}[EBL]{Electron Beam Lithography}
\acrodef{ECG}[ECG]{Electrocardiography}
\acrodef{ECoG}[ECoG]{Electrocorticography}
\acrodef{EDA}[EDA]{Electrodermal activity}
\acrodef{EDVAC}[EDVAC]{Electronic Discrete Variable Automatic Computer}
\acrodef{EEG}[EEG]{Electroencephalography}
\acrodef{EI}[EI]{Excitatory-Inhibitory}
\acrodef{EIN}[EIN]{Excitatory-Inhibitory Network}
\acrodef{EM}[EM]{Expectation Maximization}
\acrodef{EMG}[EMG]{Electromyography}
\acrodef{EOG}[EOG]{Electrooculogram}
\acrodef{EPSC}[EPSC]{Excitatory Post-Synaptic Current}
\acrodef{EPSP}[EPSP]{Excitatory Post-Synaptic Potential}
\acrodef{EZ}[EZ]{Epileptogenic Zone}
\acrodef{FAA}[FAA]{frontal alpha asymmetry}
\acrodef{FDSOI}[FDSOI]{Fully-Depleted Silicon on Insulator}
\acrodef{FET}[FET]{Field-Effect Transistor}
\acrodef{FFT}[FFT]{Fast Fourier Transform}
\acrodef{FI}[F-I]{Frequency--Current}
\acrodef{FMA}[FMA]{Floating Microelectrode Array}
\acrodef{FNN}[FNN]{Feed-forward Neural Network}
\acrodef{FPGA}[FPGA]{Field Programmable Gate Array}
\acrodef{FR}[FR]{Fast Ripple}
\acrodef{FSA}[FSA]{Finite State Automaton}
\acrodef{FSM}[FSM]{Finite State Machine}
\acrodef{GABA}[GABA]{$\gamma$-Aminobutanoic Acid}
\acrodef{GIDL}[GIDL]{Gate-Induced Drain Leakage}
\acrodef{GOPS}[GOPS]{Giga-Operations per Second}
\acrodef{GPIO}[GPIO]{General Purpose I/O}
\acrodef{GPU}[GPU]{Graphical Processing Unit}
\acrodef{GSE}[GSE]{Gaze Stationary Entropy} 
\acrodef{GT}[GT]{Ground Truth}
\acrodef{GTE}[GTE]{Gaze Transition Entropy} 
\acrodef{GUI}[GUI]{Graphical User Interface}
\acrodef{HAL}[HAL]{Hardware Abstraction Layer}
\acrodef{HFO}[HFO]{High Frequency Oscillation}
\acrodef{HH}[H\&H]{Hodgkin \& Huxley}
\acrodef{HMM}[HMM]{Hidden Markov Model}
\acrodef{HR}[HR]{Heart Rate}
\acrodef{HRS}[HRS]{High-Resistive State}
\acrodef{HSD}[HSD]{Honest Significant Difference} 
\acrodef{HSE}[HSE]{Handshaking Expansion}
\acrodef{HW}[HW]{Hardware}
\acrodef{hWTA}[hWTA]{Hard Winner-Take-All}
\acrodef{HRV}[HRV]{hearth rate variability}
\acrodef{IC}[IC]{Integrated Circuit}
\acrodef{ICA}[ICA]{Indipendent Component Analysis}
\acrodef{ICT}[ICT]{Information and Communication Technology}
\acrodef{iEEG}[iEEG]{Intracranial Electroencephalography}
\acrodef{IF2DWTA}[IF2DWTA]{Integrate \& Fire 2-Dimensional WTA}
\acrodef{IF}[I\&F]{Integrate-and-Fire}
\acrodef{IFSLWTA}[IFSLWTA]{Integrate \& Fire Stop Learning WTA}
\acrodef{IMU}[IMU]{Inertial Measurement Unit}
\acrodef{INCF}[INCF]{International Neuroinformatics Coordinating Facility}
\acrodef{INI}[INI]{Institute of Neuroinformatics}
\acrodef{IO}[I/O]{Input/Output}
\acrodef{IoT}[IoT]{Internet of Things}
\acrodef{IP}[IP]{Intellectual Property}
\acrodef{IPSC}[IPSC]{Inhibitory Post-Synaptic Current}
\acrodef{IPSP}[IPSP]{Inhibitory Post-Synaptic Potential}
\acrodef{ISI}[ISI]{Inter-Spike Interval}
\acrodef{JFLAP}[JFLAP]{Java - Formal Languages and Automata Package}
\acrodef{LEDR}[LEDR]{Level-Encoded Dual-Rail}
\acrodef{LFP}[LFP]{Local Field Potential}
\acrodef{LIFE}[LIFE]{Longitudinal Intrafascicular Electrodes}
\acrodef{LIF}[LIF]{Leaky Integrate-and-Fire}
\acrodef{LLC}[LLC]{Low Leakage Cell}
\acrodef{LMS}[LMS]{Least Mean Squares}
\acrodef{LNA}[LNA]{Low-Noise Amplifier}
\acrodef{LPF}[LPF]{Low Pass Filter}
\acrodef{LR}[LR]{Logistic Regression}
\acrodef{LRS}[LRS]{Low-Resistive State}
\acrodef{LSM}[LSM]{Liquid State Machine}
\acrodef{LTD}[LTD]{Long Term Depression}
\acrodef{LTI}[LTI]{Linear Time-Invariant}
\acrodef{LTP}[LTP]{Long Term Potentiation}
\acrodef{LTU}[LTU]{Linear Threshold Unit}
\acrodef{LUT}[LUT]{Look-Up Table}
\acrodef{LVDS}[LVDS]{Low Voltage Differential Signaling}
\acrodef{MCMC}[MCMC]{Markov-Chain Monte Carlo}
\acrodef{MAE}[MAE]{Mean Absolute Error}
\acrodef{MEA}[MEA]{Multielectrode Arrays}
\acrodef{MEMS}[MEMS]{Micro Electro Mechanical System}
\acrodef{MFR}[MFR]{Mean Firing Rate}
\acrodef{MIM}[MIM]{Metal Insulator Metal}
\acrodef{ML}[ML]{Machine Learning}
\acrodef{MLP}[MLP]{Multilayer Perceptron}
\acrodef{monoNSM}[monoNSM]{Monotonic Neural State Machine}
\acrodef{MOSCAP}[MOSCAP]{Metal Oxide Semiconductor Capacitor}
\acrodef{MOSFET}[MOSFET]{Metal Oxide Semiconductor Field-Effect Transistor}
\acrodef{MOS}[MOS]{Metal Oxide Semiconductor}
\acrodef{MRI}[MRI]{Magnetic Resonance Imaging}
\acrodef{NCS}[NCS]{Neuromorphic Cognitive Systems}
\acrodef{NDFSM}[NDFSM]{Non-deterministic Finite State Machine}
\acrodef{ND}[ND]{Noise-Driven}
\acrodef{NEF}[NEF]{Neural Engineering Framework}
\acrodef{NHML}[NHML]{Neuromorphic Hardware Mark-up Language}
\acrodef{NIL}[NIL]{Nano-Imprint Lithography}
\acrodef{NI}[NI]{Neural Interface}
\acrodef{NMDA}[NMDA]{\textit{N}-Methyl-\textsc{d}-aspartate}
\acrodef{NME}[NE]{Neuromorphic Engineering}
\acrodef{NN}[NN]{Neural Network}
\acrodef{nnNSM}[nnNSM]{Nearest Neighbors Neural State Machine}
\acrodef{NOC}[NoC]{Network-on-Chip}
\acrodef{NRZ}[NRZ]{Non-Return-to-Zero}
\acrodef{NSM}[NSM]{Neural State Machine}
\acrodef{OR}[OR]{Operating Room}
\acrodef{OTA}[OTA]{Operational Transconductance Amplifier}
\acrodef{PCB}[PCB]{Printed Circuit Board}
\acrodef{PCHB}[PCHB]{Pre-Charge Half-Buffer}
\acrodef{PCM}[PCM]{Phase Change Memory}
\acrodef{PC}[PC]{Personal Computer}
\acrodef{PDK}[PDK]{Process Design Kit}
\acrodef{PE}[PE]{Phase Encoding}
\acrodef{PFA}[PFA]{Probabilistic Finite Automaton}
\acrodef{PFC}[PFC]{Prefrontal Cortex}
\acrodef{PFM}[PFM]{Pulse Frequency Modulation}
\acrodef{PGA}[PGA]{Programmable Gain Amplifier}
\acrodef{PNI}[PNI]{Peripheral Nerve Interface}
\acrodef{PNS}[PNS]{Peripheral Nervous System}
\acrodef{PPG}[PPG]{Photoplethysmography}
\acrodef{PR}[PR]{Production Rule}
\acrodef{PSC}[PSC]{Post-Synaptic Current}
\acrodef{PSD}[PSD]{Power spectral density}
\acrodef{PSP}[PSP]{Post-Synaptic Potential}
\acrodef{PSTH}[PSTH]{Peri-Stimulus Time Histogram}
\acrodef{PV}[PV]{Parvalbumin}
\acrodef{QDI}[QDI]{Quasi Delay Insensitive}
\acrodef{RAM}[RAM]{Random Access Memory}
\acrodef{RA}[RA]{Resected Area}
\acrodef{RDF}[RDF]{Random Dopant Fluctuation}
\acrodef{RELU}[ReLu]{Rectified Linear Unit}
\acrodef{RLS}[RLS]{Recursive Least-Squares}
\acrodef{RMSE}[RMSE]{Root Mean Square-Error}
\acrodef{RRMSE}[RRMSE]{Relative Root Mean Square Error}
\acrodef{RMS}[RMS]{Root Mean Square}
\acrodef{RNN}[RNN]{Recurrent Neural Network}
\acrodef{ROLLS}[ROLLS]{Reconfigurable On-Line Learning Spiking}
\acrodef{RRAM}[R-RAM]{Resistive Random Access Memory}
\acrodef{R}[R]{Ripple}
\acrodef{RBF}[RBF]{Radial basis function}
\acrodef{RISC}[RISC]{Reduced Instruction Set Computer}
\acrodef{RSA}[RSA]{Respiratory Sinus Arrhythmia}
\acrodef{SAC}[SAC]{Selective Attention Chip}
\acrodef{SAT}[SAT]{Boolean Satisfiability Problem}
\acrodef{SCI}[SCI]{Spinal Cord Injury}
\acrodef{SCX}[SCX]{Silicon CorteX}
\acrodef{SD}[SD]{Signal-Driven}
\acrodef{SEM}[SEM]{Spike-based Expectation Maximization}
\acrodef{SCR}[SCR]{Skin Conductance Response}
\acrodef{SLAM}[SLAM]{Simultaneous Localization and Mapping}
\acrodef{SNN}[SNN]{Spiking Neural Network}
\acrodef{SNR}[SNR]{Signal to Noise Ratio}
\acrodef{SOC}[SoC]{System-On-Chip}
\acrodef{SOI}[SOI]{Silicon on Insulator}
\acrodef{SOZ}[SOZ]{Seizure Onset Zone}
\acrodef{SP}[SP]{Separation Property}
\acrodef{SPI}[SPI]{Serial Peripheral Interface}
\acrodef{SRAM}[SRAM]{Static Random Access Memory}
\acrodef{SST}[SST]{Somatostatin}
\acrodef{STDP}[STDP]{Spike-Timing Dependent Plasticity}
\acrodef{STD}[STD]{Short-Term Depression}
\acrodef{STP}[STP]{Short-Term Plasticity}
\acrodef{STT-MRAM}[STT-MRAM]{Spin-Transfer Torque Magnetic Random Access Memory}
\acrodef{STT}[STT]{Spin-Transfer Torque}
\acrodef{SVM}[SVM]{Support Vector Machine}
\acrodef{SW}[SW]{Software}
\acrodef{sWTA}[sWTA]{soft Winner-Take-All}
\acrodef{TEMP}[TEMP]{Temperature}
\acrodef{TCAM}[TCAM]{Ternary Content-Addressable Memory}
\acrodef{TFT}[TFT]{Thin Film Transistor}
\acrodef{TIME}[TIME]{Transverse Intrafascicular Multichannel Electrode}
\acrodef{TLE}[TLE]{Temporal Lobe Epilepsy}
\acrodef{UEA}[UEA]{Utah Electrode Array}
\acrodef{USB}[USB]{Universal Serial Bus}
\acrodef{USEA}[USEA]{Utah Slanted Electrode Array}
\acrodef{VHDL}[VHDL]{VHSIC Hardware Description Language}
\acrodef{VHSIC}[VHSIC]{Very High Speed Integrated Circuits}
\acrodef{VIP}[VIP]{Vasoactive Intestinal Peptide}
\acrodef{VLSI}[VLSI]{Very Large Scale Integration}
\acrodef{VNS}[VNS]{Vagal Nerve Stimulation}
\acrodef{VOR}[VOR]{Vestibulo-Ocular Reflex}
\acrodef{VSA}[VSA]{Vector Symbolic Architecture}
\acrodef{WCST}[WCST]{Wisconsin Card Sorting Test}
\acrodef{WTA}[WTA]{Winner-Take-All}
\acrodef{XML}[XML]{eXtensible Mark-up Language}

\begin{document}

\title{Neuromorphic Deployment of Spiking Neural Networks for Cognitive Load Classification in Air Traffic Control\\
\thanks{We acknowledge the financial support of the Digital Society
Initiative (DSI) at University of Zurich.}
}

\author{\IEEEauthorblockN{
Jiahui An\IEEEauthorrefmark{1}\IEEEauthorrefmark{2},
Chonghao Cai\IEEEauthorrefmark{1},
Olympia Gallou\IEEEauthorrefmark{1},
Sara Irina Fabrikant\IEEEauthorrefmark{3}
Giacomo Indiveri\IEEEauthorrefmark{1},
Elisa Donati\IEEEauthorrefmark{1}%
}
\vspace{0.2cm}

\IEEEauthorblockA{\IEEEauthorrefmark{1}Institute of Neuroinformatics, University of Zurich and ETH Zurich, Zurich, Switzerland\\ }
\IEEEauthorblockA{\IEEEauthorrefmark{2}Digital Society Initiative, University of Zurich, Zurich, Switzerland\\}

\IEEEauthorblockA{\IEEEauthorrefmark{3} Department of Geography and Digital Society Initiative, University of Zürich, Zürich, Switzerland\\ }

}

\maketitle

\begin{abstract}
This paper presents a neuromorphic system for cognitive load classification in a real-world setting, an Air Traffic Control (ATC) task, using a hardware implementation of Spiking Neural Networks (SNNs). Electroencephalogram (EEG) and eye-tracking features, extracted from an open-source dataset, were used to train and evaluate both conventional machine learning models and SNNs. Among the SNN architectures explored, a minimalistic, single-layer model trained with a biologically inspired delta-rule learning algorithm achieved competitive performance (80.6\%). To enable deployment on neuromorphic hardware, the model was quantized and implemented on the mixed-signal DYNAP-SE chip. Despite hardware constraints and analog variability, the chip-deployed SNN maintained a classification accuracy of up to 73.5\% using spike-based input. These results demonstrate the feasibility of event-driven neuromorphic systems for ultra-low-power, embedded cognitive state monitoring in dynamic real-world scenarios.
\end{abstract}


\section{Introduction}
\label{sec:intro}
Cognitive load refers to the mental effort required to process and respond to information~\cite{Sweller2011}. Accurate and real-time assessment of cognitive load is critical in domains such as human-computer interaction, education, healthcare, and aviation~\cite{Dehais_etal20,Di_etal18,Kosch_etal23,Longo_etal18}. For example, in high-stakes environments decision making contexts such as \ac{ATC}, where operators must maintain situational awareness and make rapid decisions under time pressure, inadequate cognitive load management can compromise task performance and increase the risk of errors. Therefore, reliable cognitive load monitoring is essential for developing adaptive systems that support human operators and enhance safety~\cite{Yiu_etal25}.

Conventional cognitive load inference methods rely on \ac{ML} models trained on neurophysiological signals such as \ac{EEG}~\cite{Puvsica_etal24,Di_etal18,Hassan_etal24,Ding_etal20,O_etal13}. However, their computational complexity often limits their deployment in embedded or wearable systems, where low latency and energy efficiency are critical.

Neuromorphic computing, inspired by the architecture and dynamics of biological neural systems, offers a promising alternative. \acp{SNN}, operating asynchronously and exploiting sparse spike-based communication, enable low-power, real-time processing~\cite{Chicca_etal14,O_etal13}. In this context, neuromorphic processors such as the \ac{DYNAP-SE} chip~\cite{Moradi_etal18} provide an efficient hardware substrate for continuous cognitive state monitoring. The \ac{DYNAP-SE} features four cores, each comprising 256 \ac{ADEXP} neurons with configurable excitatory and inhibitory synaptic dynamics, and uses an asynchronous \ac{AER} communication protocol for scalable inter-core connectivity.

Previous work has demonstrated the utility of the \ac{DYNAP-SE} chip in various biosignal processing tasks, including anomaly detection in \ac{ECG} signals~\cite{Bauer_etal19,Carpegna_etal24}, classification of \ac{PPG} waveforms~\cite{DeLuca_etal25}, recognition of \ac{EMG} patterns~\cite{Donati_etal19,Ma_etal20a,Ceolini_etal20}, and biomarker extraction from \ac{EEG} recordings in epilepsy patients~\cite{Sharifshazileh_etal21,Gallou_etal24}. These studies highlight the potential of \acp{SNN} as biologically inspired alternatives to conventional \ac{ML} approaches, including in domains such as cognitive load monitoring.

\begin{figure*}[ht]
    \centering

    \begin{subfigure}[b]{0.31\textwidth}
        \centering
        \includegraphics[width=\linewidth]{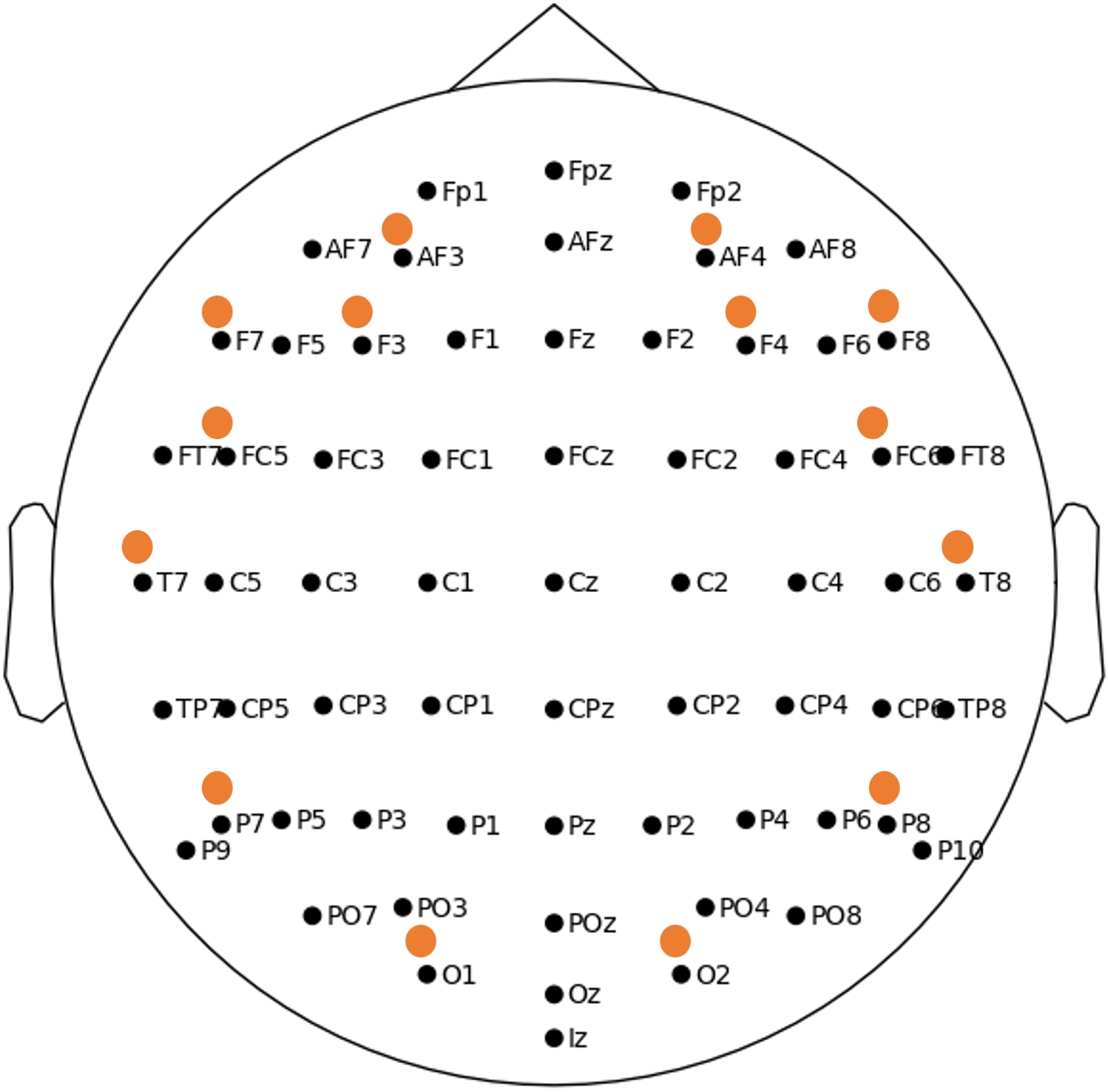}
        \caption*{(a)}
        \label{fig:sub_10_20}
    \end{subfigure}
    \hfill
    \begin{subfigure}[b]{0.35\textwidth}
        \centering
        \includegraphics[width=\linewidth]{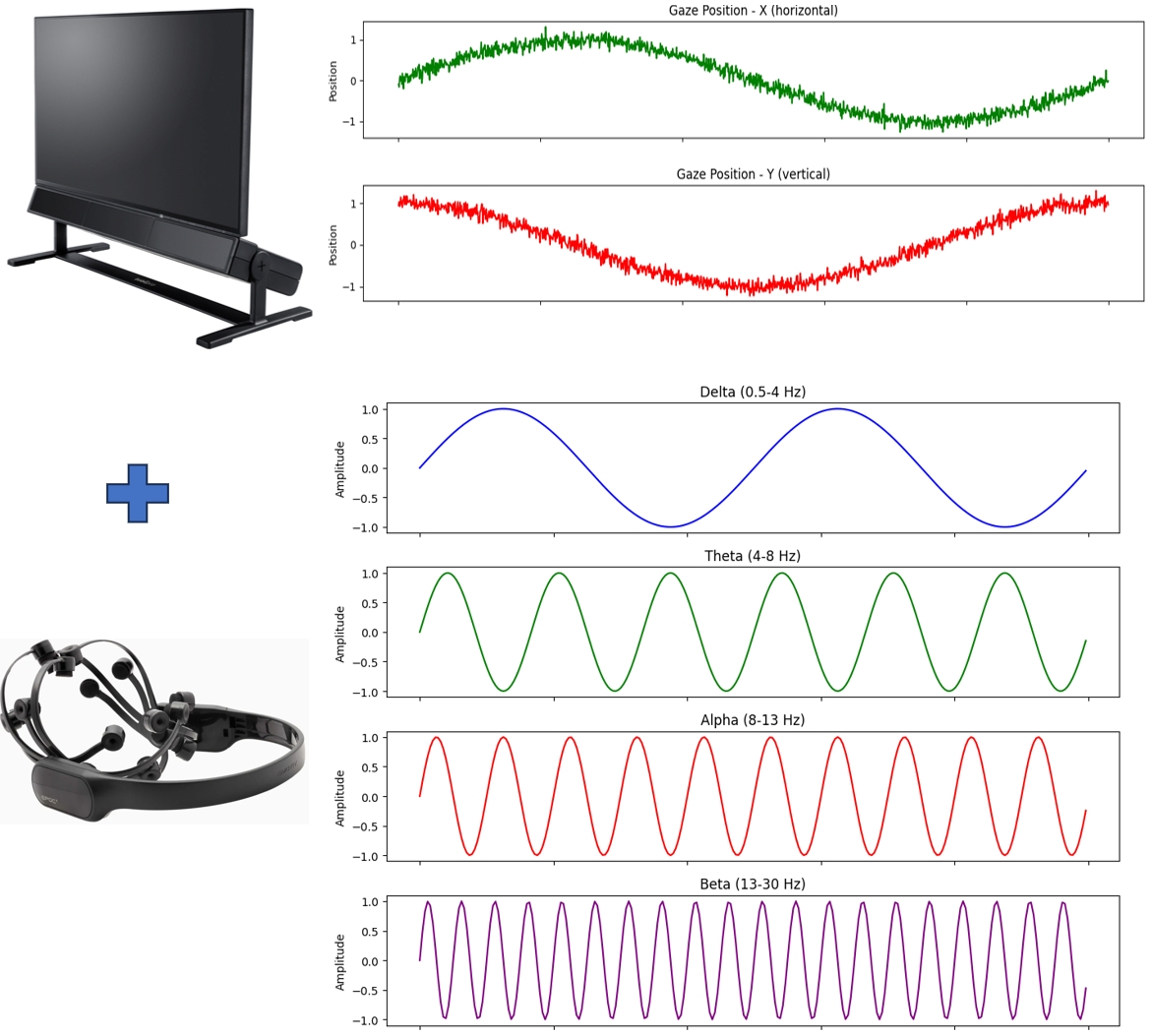}
        \caption*{(b)}
        \label{fig:sub_eeg_eyetracker}
    \end{subfigure}
    \hfill
    \begin{subfigure}[b]{0.31\textwidth}
        \centering
        \includegraphics[width=\linewidth]{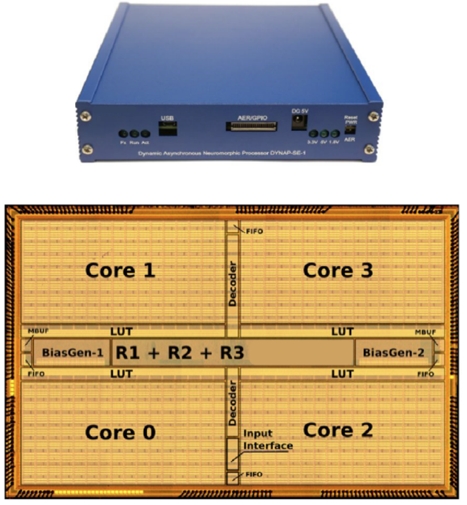}
        \caption*{(c)}
        \label{fig:sub_dynapse}
    \end{subfigure}

    \caption{System overview of the multimodal cognitive load classification setup. (a) The EEG electrode layout follows the standard 10–20 system, with Emotiv EEG sensor positions highlighted in orange. (b) EEG signals were recorded using Emotiv EPOC+ EEG headset, while gaze data were captured via a screen-based eye tracker. The panel also shows representative EEG and gaze-derived physiological features, including gaze position (X, Y), and oscillatory brain activity across standard frequency bands. (c) These features were used as input to a neuromorphic model. The extracted features were processed using the mixed-signal spiking neural network processor DYNAP-SE, which contains four neuromorphic cores, each supporting up to 256 analog neurons.}
    \label{fig:system_overview}
\end{figure*}

In this work, we present a neuromorphic system that classifies cognitive load during an \ac{ATC} task, using multimodal physiological features extracted from an ecologically valid laboratory experiment involving both professional air traffic controllers and novices trained on the task~\cite{LaniniMaggi_etal21}. We train and evaluate both conventional \ac{ML} models and \acp{SNN}, with the latter implemented in software (snnTorch~\cite{Eshraghian_etal23}) and deployed on the mixed-signal \ac{DYNAP-SE} hardware platform.

Among the evaluated \ac{SNN} architectures, a minimalistic no-hidden-layer model trained using a biologically inspired delta-rule learning algorithm achieved competitive performance (80.6\%) and, after quantization, maintained robust classification accuracy (up to 73.5\%) when deployed on hardware.
Since the \ac{DYNAP-SE} chip does not support online learning, the \ac{SNN} was first trained offline in simulation and subsequently deployed onto the hardware with fixed weights for inference.

To the best of our knowledge, this work presents the first complete demonstration of real-time cognitive state classification from \ac{EEG} and eye-tracking signals using quantized \acp{SNN} deployed on neuromorphic hardware. Figure~\ref{fig:system_overview} summarizes the multimodal data acquisition, feature extraction, and spike-based neuromorphic processing pipeline that enables energy-efficient cognitive load monitoring.

\section{Materials \& Methods}
\label{sec:methods}
We designed a complete neuromorphic cognitive load classification pipeline, encompassing dataset selection and preprocessing, feature extraction from multimodal physiological signals, model training with both conventional \ac{ML} algorithms and \acp{SNN}, and hardware deployment on the mixed-signal \ac{DYNAP-SE} platform. The following sections detail each stage of the methodology.

\subsection{Dataset and Experimental Paradigm}
\label{ssec:dataset}
We used an open-access dataset from an ecologically validated \ac{ATC} experiment~\cite{LaniniMaggi_etal21}, in which both professional air traffic controllers and novice participants monitored animated aircraft displays to detect potential collisions. \ac{EEG} and eye-tracking data were simultaneously recorded during each 4-second trial under varying task difficulty (4 vs. 8 animated objects). \ac{EEG} signals were collected using a 14-channel (AF3, F7, F3, FC5, T7, P7, O1, O2, P8, T8, FC6, F4, F8, AF4) Emotiv EPOC+ device\cite{Emotiv2021} and gaze data were recorded with a Tobii TX300 eye tracker. The full dataset includes 592 trials from 39 participants. For this study, we retained 304 trials from 19 participants who had at least one valid modality (\ac{EEG} or eye-tracking), including 9 professional air traffic controllers and 10 control participants. A total of 217 trials containing both \ac{EEG} and gaze-derived features were used for model training and hardware deployment. Preprocessing and feature extraction followed the protocol described in~\cite{LaniniMaggi_etal21}.

Five high-level features were extracted as model inputs: $\alpha$ power, \ac{EEG} engagement index, \ac{FAA}, \ac{GTE}, and \ac{GSE}. These features captured oscillatory neural dynamics and visual exploratory behavior. Specifically, $\alpha$ power was computed as the mean spectral power in the 8–12 Hz band across selected channels; the \ac{EEG} engagement index was defined as $\beta/(\alpha + \theta)$, reflecting attentional and arousal states; and \ac{FAA} was calculated as the log ratio of right to left frontal $\alpha$ power, indexing affective and cognitive regulation.
Complementary features from eye-tracking included \ac{GTE}, quantifying the randomness of gaze transitions between predefined \acp{AOI}, and \ac{GSE}, measuring the spatial dispersion of fixations across \acp{AOI}.

Figure~\ref{fig:system_overview} summarizes the multimodal acquisition and processing pipeline. \ac{EEG} and gaze signals were collected using wearable and screen-based devices, respectively, and processed to extract frequency-specific activities (delta, theta, alpha, beta, gamma), gaze positions (X, Y), and more high-level cognitive features. These features were then encoded into spike-based representations for classification using a neuromorphic model deployed on the \ac{DYNAP-SE} board, a mixed-signal processor comprising four cores with 256 analog neurons each.

Task difficulty was operationalized as a binary classification label, defined by the number of aircraft displayed (4 = easy, 8 = hard), and encoded as 0 (low) and 1 (high) to maintain a balanced dataset.

\subsection{Model Training and Evaluation}
\label{ssec:model}
We implemented and evaluated several classification models to distinguish between different levels of cognitive load based on five physiological features derived from \ac{EEG} and eye-tracking signals. These features included alpha power, \ac{EEG} engagement index, \ac{FAA}, \ac{GTE}, and \ac{GSE}. All features were standardized with z-score normalization using the \texttt{StandardScaler} function from scikit-learn before being input into the models.

Model performance was assessed using four key metrics: accuracy, F1-score, precision, and recall. These metrics were computed on an independent 20\% testing set. To account for variability across participants, all performance metrics were averaged across stratified $k$-fold cross-validation splits ($k = 5$).

\subsubsection{Machine Learning models as Baselines}
\label{ssec:training}
As a baseline, we implemented standard models commonly used in the literature for cognitive load decoding. Establishing these baselines allows us to better characterize the dataset and provides a reference point to evaluate the performance of the proposed \ac{SNN} models.

We first implemented a \ac{LR} model with L2 regularization. A grid search was conducted over the inverse regularization parameter \( C \in \{0.01, 0.1, 1, 10, 100\} \) using a 5-fold stratified cross-validation strategy to find the optimal configuration. In parallel, we trained \ac{SVM} classifiers with three different kernels: linear, polynomial, and \ac{RBF}. For each kernel, a separate hyperparameter tuning process was employed. For the linear kernel, only the regularization strength \( C \) was tuned, whereas for the polynomial and \ac{RBF} kernels, kernel-specific parameters (degree and gamma) and class weights were also optimized. All models were tuned using grid search with 5-fold stratified cross-validation.

\subsubsection{Training \ac{SNN} with Delta Learning Rule}
\label{sssec:training}
To explore the potential of event-based models for cognitive load decoding, we tested different \ac{SNN} architectures using the \texttt{snntorch} library~\cite{Eshraghian_etal23}, incorporating chip constraints to enable future deployment on neuromorphic hardware. The input to each \ac{SNN} was a temporally encoded spike train generated from the five feature values using a rate-based \ac{LIF} encoder. The \ac{LIF} model follows the standard differential equation governing membrane voltage dynamics:
\begin{equation}
    \tau \frac{dV(t)}{dt} = -(V(t) - V_{\text{rest}}) + X(t)
    \label{eq:lif}
\end{equation}
where \( V(t) \) is the membrane potential at time \( t \), \( V_{\text{rest}} \) is the resting potential (set to 0), \( \tau \) is the membrane time constant, and \( X(t) \) represents the input feature vector. A spike is emitted when \( V(t) \) exceeds the firing threshold \( V_{\text{th}} = -50 \), after which the potential is reset to \( V_{\text{reset}} = -65 \).

The temporal encoder was run for 16 simulation steps, corresponding to the number of trials per subject, with each trial representing 4 seconds of cognitive engagement. The encoded output was a spike count vector for each feature across the 16 steps. These binary spike trains served as input to the \ac{SNN}.

The network architecture comprised two fully connected layers: the first layer (fc1) projected the inputs to a hidden spiking layer via non-plastic, sparsely connected weights; the second layer (fc2) applied a delta learning rule based on the mismatch between predicted and target spike activity. The output spiking layer consisted of two \ac{LIF} neurons with weak mutual inhibition to enable \ac{WTA} dynamics. The inhibition matrix was predefined as:
\[
\begin{bmatrix}
0 & -\gamma \\
-\gamma & 0
\end{bmatrix}
\]
where \( \gamma \in [0.01, 0.05, 0.1] \) controlled the strength of mutual inhibition between competing output neurons.

The fc1 weights were initialized from a normal distribution with a tunable mean (\textit{fc1\_mean}) and a fixed standard deviation of \( 0.2 \times \textit{fc1\_mean} \). A binary connection mask was applied according to a tunable connection probability parameter \( P_{\text{conn}} \), enabling the modeling of biologically plausible sparse connectivity. All fc1 weights remained frozen during training, while fc2 weights were updated using the delta rule. The output \ac{LIF} neurons also had learnable \(\beta\) parameters to adapt their membrane decay dynamics, tuned via grid search.

The output activity \( \hat{y} \in \mathbb{R}^{2} \) was computed as the \ac{LIF}-encoded spike response after applying inhibition. Classification was performed by selecting the neuron with the highest firing rate. The delta rule weight update was defined as:
\begin{equation}
\Delta w = \eta \cdot (y_{\text{true}} - y_{\text{pred}}) \cdot x
\end{equation}
where \( \eta \) is the learning rate and \( x \) is the presynaptic activation from fc1.

A grid search was performed to optimize biologically relevant hyperparameters, including the membrane time constant \( \tau \in \{15, 20, 25, 30, 35, 40\} \), the fc1\_mean \( \in \{0.1, 0.3, 0.5, 0.7\} \), the connection probability \( P_{\text{conn}} \in \{0.2, 0.5, 0.8\} \), and the inhibition strength \( \gamma \in \{0.01, 0.05, 0.1\} \). Each configuration was trained for 20 epochs using a learning rate of 0.001 and a batch size of 1. All experiments were repeated across three random seeds (\{0, 21, 42\}) to ensure robustness.

\paragraph{\ac{SNN} with Hidden Layer}
To evaluate the impact of network depth on classification performance, we implement \ac{SNN} architectures by introducing a hidden spiking layer. Adding a hidden layer increases the model's capacity to capture nonlinear feature interactions, enabling the system to learn more complex decision boundaries while maintaining sparse and biologically plausible connectivity. Specifically, we varied the hidden layer size as a multiple of the input dimension (\(5\)), testing three scales: \(3\times\), \(5\times\), and \(10\times\).

\begin{figure}[ht]
    \centering
    \begin{subfigure}[b]{0.48\linewidth}
        \centering
        \includegraphics[width=\linewidth,height=6.9cm,keepaspectratio]{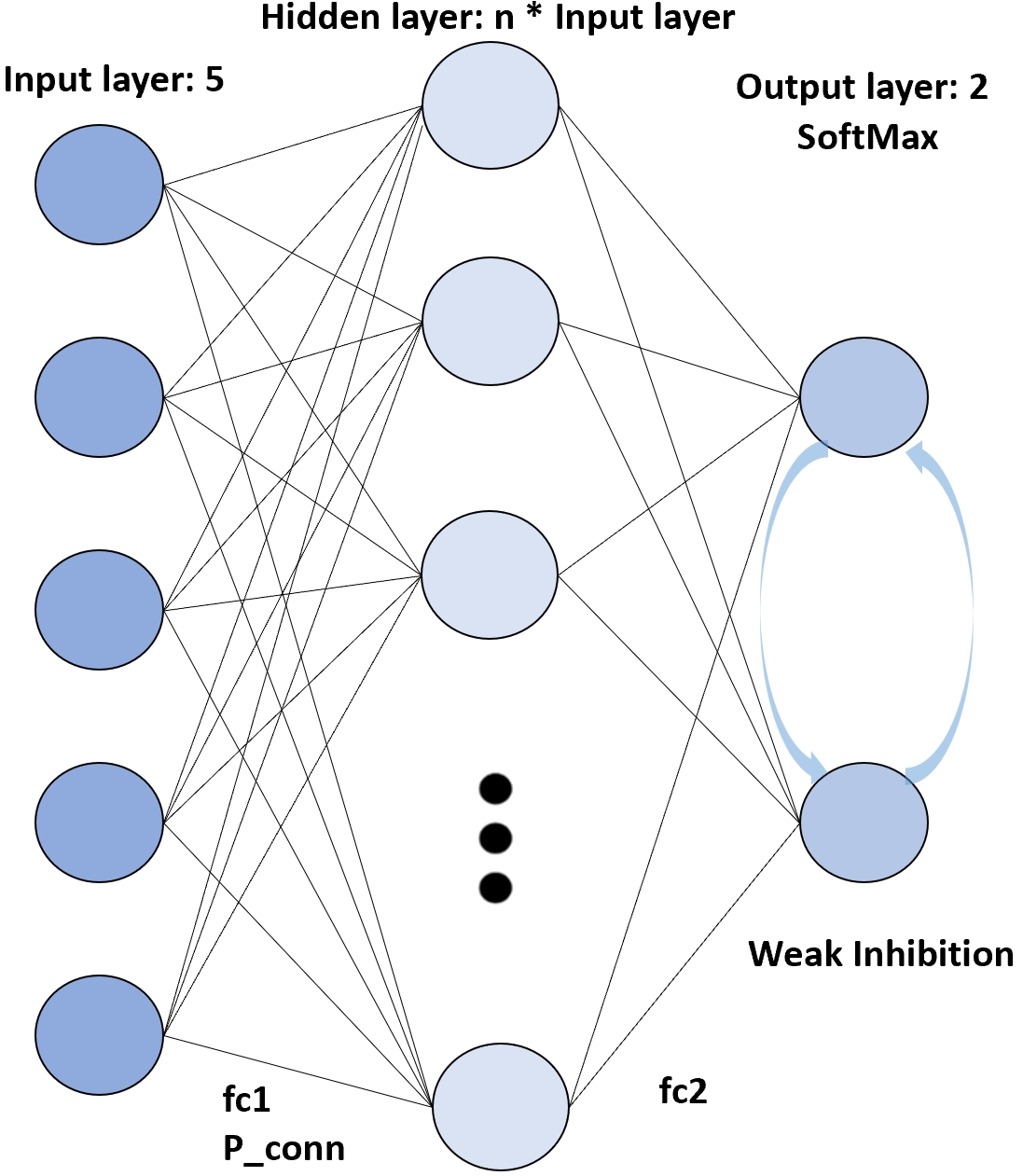}
        \caption{\ac{SNN} with hidden layer}
    \end{subfigure}
    \hfill
    \begin{subfigure}[b]{0.48\linewidth}
        \centering
        \includegraphics[width=\linewidth,height=5.0cm,keepaspectratio]{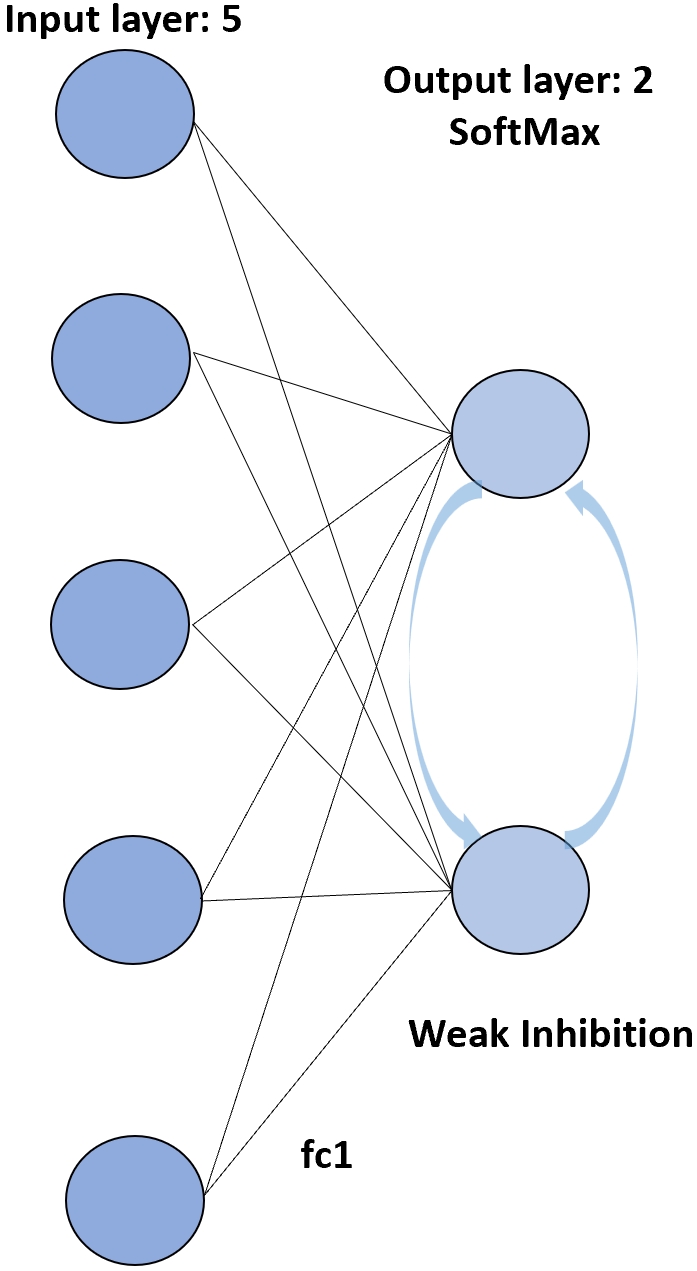}
        \caption{\ac{SNN} without hidden layer}
    \end{subfigure}
    \caption{Architectures of the \acp{SNN} used for cognitive load classification. (a) \ac{SNN} with hidden layer: Input features are spike-encoded using \ac{LIF} neurons and passed through a sparse (P\_conn) fixed-weight layer (fc1), followed by a trainable output layer (fc2). (b) SNN without hidden layer: Spike-encoded inputs are directly connected to the output layer (fc1), with no intermediate hidden neurons. In both models, the output layer consists of two \ac{LIF} neurons with weak mutual inhibition and a softmax readout, enabling \ac{WTA} behavior. All output weights are trained using a delta learning rule adapted to spiking activity.}
    \label{fig:snn_architecture_bio}
\end{figure}

A visual overview of this biologically grounded architecture is shown in Fig.~\ref{fig:snn_architecture_bio}(a).

\paragraph{\ac{SNN} without Hidden Layer}
Based on the \ac{ML} results, we also implemented a single-layer \ac{SNN} to evaluate the performance of a minimal architecture with primarily linear separation capabilities, as shown in Fig.~\ref{fig:snn_architecture_bio}(b). In this version, the spike-encoded inputs are directly connected to two output neurons modeled as \ac{LIF} neurons, enabling spike-based representation and computation without intermediate hidden processing.

The input features were first encoded using the same spike encoding strategy described previously. The spiking classifier consisted of a single fully connected linear layer with two output neurons, followed by a \ac{LIF} layer with a learnable decay factor \( \beta \) and a weak lateral inhibition term between the two outputs, as in the previous \ac{SNN} models. Training was performed using a delta rule adapted to spiking output, where the weights of the input-to-output connections were iteratively updated based on the difference between the target label and the softmax-transformed spike output.

An extensive grid search was conducted to identify the best-performing model configuration, optimizing the membrane time constant \( \tau \in \{15,18,20,23,25,28,30,31,32,33,35\} \), the learning rate \( \eta \in \{0.0001, 0.001, 0.01\} \), the number of training epochs \( \in \{20, 25, 30\} \), and the inhibition strength \( \gamma \in \{0.01, 0.05, 0.1\} \). Each experiment was repeated across 12 random seeds to ensure robustness, and stratified 5-fold cross-validation was used to assess generalization performance.

\subsection{Deployment of Quantized SNN on DYNAP-SE chip}
\label{ssec:deployment_method}
To evaluate real-world applicability, we deployed the optimal single-layer \ac{SNN} on the neuromorphic chip \ac{DYNAP-SE}. Since the \ac{DYNAP-SE} chip does not support online learning, the network was first trained offline and subsequently deployed with fixed weights for inference. The deployment involved a multi-step process comprising spike encoding, weight quantization, network mapping, and population-based spike decoding. The model selected for deployment corresponded to the best-performing fold identified during 5-fold cross-validation of the software simulation, achieving the highest performance among all folds.

\paragraph{Spike Conversion and Encoding}
First, the continuous input features were encoded into spike trains using the same \ac{LIF} dynamics described earlier (with grid-searched optimal \( \tau \) and 16 simulation steps). Each trial sample produced a spike raster with up to five input units corresponding to the multimodal features. To prepare the input for \ac{DYNAP-SE}-compatible spike generators, we converted the spike rasters into timestamped firing events (in milliseconds). For each trial, binary spike matrices were generated, where a value of 1 indicated the occurrence of a spike for a particular feature at a given time step. These matrices were then transformed into event-based spike dictionaries containing spike times and corresponding unit indices.

To meet \ac{DYNAP-SE} input constraints, a linear time scaling and offset shift were applied to each spike's time step. Specifically, the spike time \( t \) was calculated as:
\[
t = (\text{time step} + 1) \times 0.01 \, \text{seconds}.
\]
Only trials with complete spike activity across all time steps were retained to ensure consistency across the dataset.

\paragraph{Quantization of Network Weights}
To meet the \ac{DYNAP-SE} chip’s precision constraints (supporting only 3- or 4-bit signed weights), we applied 3-bit (int3) quantization to the learned float weights and biases from the best-performing fold. Quantization was performed by linearly scaling the float weights to the range \([-3, 3]\), rounding to the nearest integer, and clipping any values exceeding the representable bounds.

The resulting quantized parameters were loaded into a new \ac{SNN} model with fixed weights and retrained using the same delta rule to assess the impact of quantization. The spike-based input encoding and \ac{LIF} neuron dynamics were preserved to ensure compatibility with the event-driven architecture of the \ac{DYNAP-SE} chip.

\paragraph{Hardware Network Mapping}
The network deployed on the \ac{DYNAP-SE} chip included five input spike generators (representing the five physiological features) and two output populations, each consisting of 20 neurons mapped to separate cores (Core 1 and Core 2). Spike generators were used to inject timestamped input events into the chip, while each output population acted as a readout unit corresponding to one of the two cognitive load classes.

Each feature-to-output population connection was weighted using an integer-valued \( 5 \times 2 \) weight matrix with values ranging from \(-3\) to \(+3\). To implement these weights in hardware:
\begin{itemize}
    \item Positive weights were realized through multiple parallel synaptic connections of type AMPA (fast excitatory).
    \item Negative weights were realized through multiple parallel synaptic connections of type GABA-A (fast inhibitory).
    \item The number of parallel synapses matched the absolute value of the weight; for example, a weight of \(+3\) created three AMPA synapses.
\end{itemize}

The network was assembled using the \ac{DYNAP-SE} API. Neurons in each output population received input from spike generators according to the quantized weight matrix and were further interconnected through GABA-B slow inhibitory synapses, implementing lateral inhibition between the two output populations to enable competitive decision dynamics.

\paragraph{Monitoring and Evaluation}
Each encoded trial was presented to the chip via the FPGA-based spike generator interface. For each trial, the simulation ran for 0.16 seconds, during which output spikes were recorded from the two output populations. To ensure system stability, a buffer period of 0.5 seconds was added before and after stimulation to allow neuron dynamics to relax.

The recorded spike activity was post-processed by binning the spikes into discrete time windows:
\begin{itemize}
    \item A window size of 2.5 ms (0.0025 s) was used.
    \item For each window, the average firing rate of each output population was computed by dividing the number of spikes by the product of the window length and the number of neurons (20 per population).
\end{itemize}

Classification was performed using a burst-based decoding strategy. Bursting events were identified by detecting windows with elevated firing rates, and the predicted class was assigned based on which output population exhibited stronger or more frequent bursts. In cases of ambiguity, a dynamic threshold adjustment was applied to resolve ties. Fallback rules ensured a decision even under low-activity conditions.

\begin{algorithm}
\footnotesize
\caption{Burst-Based Classification}
\begin{algorithmic}
\STATE \textbf{Input:} Population firing rates; thresholds: \textit{zero}, \textit{diff}, \textit{offset}, \textit{limit}
\STATE \textbf{Output:} Predicted class \textit{iclass}
\STATE Compute max\_rate\_pop1 and max\_rate\_pop2
\IF{both max rates $>$ \textit{zero}}
    \IF{rate difference $\geq$ \textit{diff}}
        \STATE \textit{iclass} $\gets$ population with higher max rate
    \ELSE
        \STATE Count windows where rate $\geq$ (max rate $-$ \textit{offset})
        \WHILE{counts equal \AND offset $<$ \textit{limit}}
            \STATE Increase offset; re-count
        \ENDWHILE
        \STATE \textit{iclass} $\gets$ population with more counts
    \ENDIF
\ELSIF{only one max rate $>$ \textit{zero}}
    \STATE \textit{iclass} $\gets$ that population
\ELSE
    \STATE \textit{iclass} $\gets$ population with more total spikes
\ENDIF
\end{algorithmic}
\end{algorithm}

To assess the robustness and stability of the deployed \ac{SNN} on the \ac{DYNAP-SE} chip, we conducted five independent inference trials using the same optimized configuration. Evaluation metrics, including accuracy, precision, recall, and F1-score, were computed separately for each trial to capture variability arising from analog hardware stochasticity and spike dynamics.

\section{Results}
\label{sec:results}

\subsection{Machine Learning Model Performance}
\label{ssec:ml_results}
We evaluated the classification performance of several conventional \ac{ML} models, including \ac{LR} and \acp{SVM} with linear, polynomial, and \ac{RBF} kernels.

Across all classifiers, the integration of multimodal cognitive features from \ac{EEG} and Eyetracker—led to consistent performance improvements. Among these, the best-performing model was the polynomial-kernel \ac{SVM}, achieving an accuracy of 84.3\% ($\pm$ 5.6\%), F1-score of 84.2\% ($\pm$ 5.6\%), and precision of 84.8\% ($\pm$ 5.9\%). The \ac{RBF} \ac{SVM} followed closely with an accuracy of 83.4\% ($\pm$ 6.9\%), while \ac{LR} also demonstrated strong generalization with 82.5\% ($\pm$ 6.6\%) accuracy and 82.3\% ($\pm$ 6.7\%) F1-score.

The detailed results for each classifier are presented in Table~\ref{tab:ml_performance}.
\begin{table}[ht]
\caption{Performance Comparison of ML Models}
\label{tab:ml_performance}
\centering
\renewcommand{\arraystretch}{1.2}
\begin{tabular}{|l|c|c|c|c|}
\hline
\textbf{Model} & \textbf{Accuracy} & \textbf{F1} & \textbf{Precision} & \textbf{Recall} \\
\hline
LR & 82.5\% & 82.3\% & 82.8\% & 82.3\% \\
SVM (Linear) & 82.9\% & 82.6\% & 83.9\% & 82.6\% \\
SVM (Poly) & 84.3\% & 84.2\% & 84.8\%& 84.2\% \\
SVM (RBF) & 83.4\% & 83.2\% & 84.1\% & 83.3\% \\
\hline
\end{tabular}
\end{table}

\subsection{Spiking Neural Network Performance}
\label{ssec:snn_results}
We explored \acp{SNN} for cognitive load classification using spike-encoded physiological features.
Continuous input values were encoded into spike trains using a \ac{LIF} neuron model, configured with a resting potential \( V_{\text{rest}} = 0 \), reset potential \( V_{\text{reset}} = -65.0 \), and firing threshold \( V_{\text{th}} = -50.0 \), over a simulation window of 16 steps.
The membrane time constant \( \tau \) was treated as a hyperparameter and optimized via grid search.

\paragraph{\ac{SNN} with Hidden Layers}
To test the effect of deeper biological models, we implemented biologically plausible \ac{SNN} models using fixed weights for fc1, sampled from a Gaussian distribution with a given mean (\textit{fc1\_mean}) and connection probability (\textit{P\_conn}). Only fc2 weights were trained using the delta learning rule. The membrane dynamics were governed by the standard \ac{LIF} differential equation, and hyperparameters were optimized using a nested grid search. The grid search tuned \( \tau \), \textit{fc1\_mean}, \textit{P\_conn}, inhibition strength \( \gamma \), and random seeds across various hidden layer sizes set to \(3\times\), \(5\times\), and \(10\times\) the input size. Each configuration was evaluated using 5-fold cross-validation.

The best \(3\times\) hidden \ac{SNN} achieved an accuracy of \(65.4\% \pm 6.8\%\), with a high precision (\(79.2\%\)) but lower recall (\(51.3\%\)), suggesting cautious prediction behavior.
The \(5\times\) hidden configuration improved recall (\(90.4\%\)) and F1-score (\(74.9\%\)) at \(67.3\%\) accuracy.
The \(10\times\) configuration yielded the best F1-score of \(68.3\%\), balancing both precision (\(76.5\%\)) and recall (\(53.0\%\)).

The best-performing model (\(10\times\) hidden, seed = 16, \( \tau=30 \), \textit{fc1\_mean} = 0.4, \( \textit{P\_conn}=0.5 \), inhibition= \(0.01\)) achieved \(79.2\% \pm 4.9\%\) accuracy and an F1-score of \(80.2\% \pm 3.2\%\), surpassing the baseline \ac{LR} model and approaching the performance of kernelized \ac{SVM} models. These results demonstrate that biological constraints, such as sparse fixed input weights and lateral inhibition, can lead to strong classification performance when appropriately tuned.

\paragraph{Single-layer \ac{SNN}}
To explore minimal neuromorphic implementations, we additionally trained a single-layer \ac{SNN}. This architecture employed \ac{LIF} spike encoding and a biologically inspired delta-rule learning mechanism applied directly to the output layer. Key hyperparameters—including the membrane time constant (\( \tau \)), learning rate (\( \eta \)), inhibition strength, and number of training epochs—were tuned through grid search across multiple random seeds.

The best configuration (seed = 31, \( \tau = 31 \), inhibition = 0.01, \( \eta=0.01 \), 25 epochs) achieved an accuracy of \(80.6\% \pm 4.9\%\), precision of \(85.2\% \pm 5.1\%\), recall of \(77.4\% \pm 8.1\%\), and F1-score of \(80.6\% \pm 1.6\%\). Despite its simplicity, this architecture performed competitively, highlighting its potential for efficient, hardware-constrained deployment.

Table~\ref{tab:snn_summary} compares the performance of this minimal \ac{SNN} with its multi-layer counterparts, trained with varying hidden layer sizes (\(3\times\), \(5\times\), and \(10\times\) the input dimension).
\begin{table}[ht]
\caption{Performance Comparison of \acp{SNN} with and without Hidden Layer}
\label{tab:snn_summary}
\centering
\renewcommand{\arraystretch}{1.2}
\begin{tabular}{|l|c|c|c|c|}
\hline
\textbf{Model} & \textbf{Accuracy} & \textbf{F1 Score} & \textbf{Precision} & \textbf{Recall} \\
\hline
No Hidden    & 80.6\%  & 80.6\%  & 85.2\%  & 77.4\% \\
\hline
3× Hidden     & 65.3\%   & 60.0\% & 79.2\% & 51.3\% \\
\hline
5× Hidden     & 67.3\%  & 74.9\%  & 64.8\%  & 90.4\% \\
\hline
10× Hidden    & 72.9\%   & 68.3\% & 85.6\% & 57.4\% \\
\hline
\end{tabular}
\end{table}

\subsection{Quantized \ac{SNN} Deployment Results}
\label{ssec:deployment_results}
We evaluated the quantized \ac{SNN} using stratified 5-fold cross-validation. To meet the precision constraints of the \ac{DYNAP-SE} chip, int3 quantization was applied to the float weights learned from the best-performing fold. The resulting weight matrix was:
\[
\text{Weights}_{\text{int3}} =
\begin{bmatrix}
0 & -1 & -1 & 3 & -2 \\
0 & 1 & 1 & -3 & 2
\end{bmatrix}
\]

These quantized parameters were loaded into a new \ac{SNN} model with fixed weights and retrained using the same delta learning rule to assess the impact of quantization. Spike-based input encoding and \ac{LIF} neuron dynamics were preserved to maintain event-driven compatibility.

Despite reduced weight precision, the quantized model maintained strong performance, achieving an average accuracy of \(76.9\%\), precision of \(70.8\%\), recall of \(96.5\%\), and F1-score of \(81.7\%\). Compared to the original floating-point model, the accuracy dropped by only 3.7 percentage points, while high recall indicated robust sensitivity to the target class.

The quantized \ac{SNN} was subsequently deployed on the \ac{DYNAP-SE} hardware and evaluated on 217 valid samples (47\% easy, 53\% hard labels). Classification was performed fully on-chip by feeding spike train inputs and monitoring the differential response of output neuron populations mapped to separate chip cores. To assess robustness under hardware variability, five independent inference trials were conducted using the same configuration. Across trials, the model achieved an average accuracy of \(73.5\% \pm 0.6\%\), precision of \(70.4\% \pm 0.3\%\), recall of \(86.1\% \pm 1.4\%\), and F1-score of \(77.5\% \pm 0.6\%\).

\begin{figure}[ht]
    \centering
    \includegraphics[width=\linewidth]{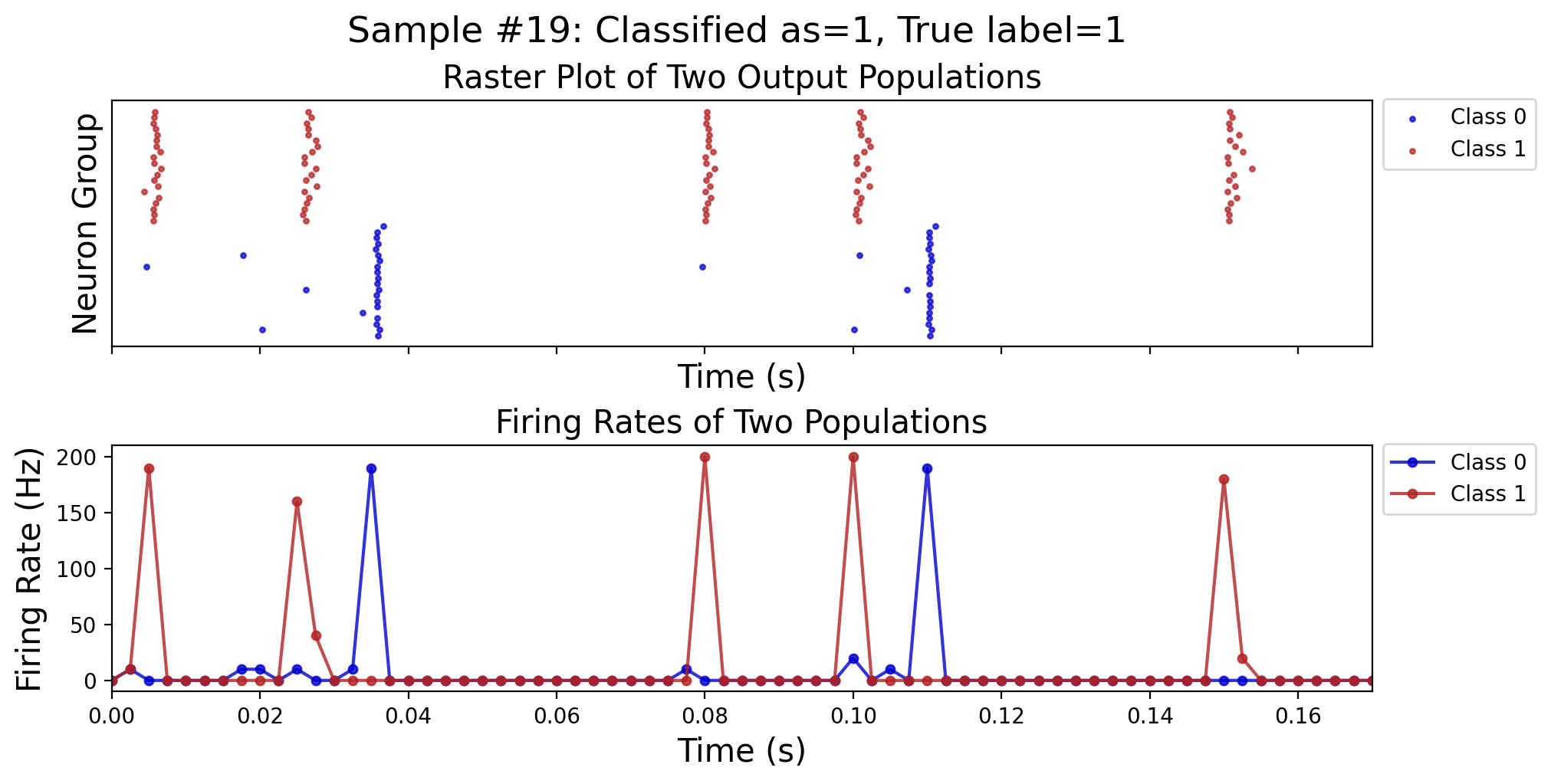}
    \caption{Example Dynapse trial result illustrating the spike-based classification process. The raster plot (top) shows spike events from output neurons of Class 0 (blue) and Class 1 (red). The bottom panel shows sliding-window firing rates computed over 2.5 ms intervals. In this trial, stronger activity from the Class 1 population led to a correct Class 1 classification.}
    \label{fig:dynapse_example_result}
\end{figure}

Figure~\ref{fig:dynapse_example_result} shows a representative trial output, illustrating the differential spiking activity and firing rate dynamics that led to a correct Class 1 prediction.

These results demonstrate that, despite quantization and hardware constraints, the deployed \ac{SNN} maintained strong classification performance without retraining, confirming the feasibility of ultra-low-power, wearable neuromorphic systems for continuous cognitive state monitoring. However, the observed variability across hardware trials highlights the intrinsic stochasticity of analog neuromorphic platforms such as \ac{DYNAP-SE}. Even with fixed weights and identical spike inputs, variations in spike timing, neuron dynamics, or lateral inhibition interactions can significantly affect network behavior. Robust deployment thus requires careful tuning of both network architecture and runtime parameters to mitigate hardware-induced variability.

\section{Discussion}
\label{sec:discussion}
Overall, our findings demonstrate that a simple, single-layer \ac{SNN}, trained using biologically plausible learning rules, can be effectively quantized and deployed on neuromorphic hardware with state-of-the-art classification performance.

\subsection{Effectiveness of Multimodal Cognitive Features}
\label{ssec:multimodal_discussion}

Across all models, multimodal features derived from \ac{EEG} and eye-tracking data yielded robust classification performance. These results confirm that combining oscillatory neural activity and visual entropy metrics enhances cognitive load decoding, particularly in dynamic \ac{ATC} tasks.

\subsection{Insights into Linearity and Model Complexity}
\label{ssec:ml_discussion}
All conventional \ac{ML} models, including \ac{LR} and \ac{SVM} variants, achieved similar accuracy levels (82.5\%–84.3\%, Table~\ref{tab:ml_performance}). This narrow performance gap suggests that the task is largely linearly separable in the selected feature space. Notably, the \ac{LR} model, despite its simplicity, achieved 82.5\% accuracy, indicating that model interpretability and low complexity can be achieved without compromising performance.

\subsection{Simplicity and Robustness of single-layer \ac{SNN}}
\label{ssec:snn_no_hidden}
The single-layer \ac{SNN} trained with delta-rule learning reached 80.6\% accuracy and 85.2\% precision (Table~\ref{tab:snn_summary}). Its strong performance despite minimal architecture highlights the adequacy of the selected features and the potential of low-capacity spiking networks for embedded neuromorphic applications.

\subsection{Impact of Hidden Layer Size in \ac{SNN}}
\label{ssec:snn_hidden}
Expanding the hidden layer size to 3$\times$, 5$\times$, and 10$\times$ the input dimension did not consistently improve performance. The 5$\times$ hidden model achieved the highest F1-score (74.9\%) and recall (90.4\%), while the 10$\times$ model optimized accuracy (72.9\%) and precision (85.6\%). Overfitting tendencies were observed in wider models, while moderate-width networks provided a better balance between sensitivity and stability. Interestingly, the single-layer \ac{SNN} outperformed the 3$\times$ and 5$\times$ hidden variants, reinforcing the strength of the feature set and the utility of shallow spike-based architectures for this task.

\subsection{Hardware Deployment and Variability in \ac{DYNAP-SE}}
\label{ssec:deployment_discussion}
The quantized single-layer \ac{SNN} was successfully mapped to the \ac{DYNAP-SE} chip and evaluated on 217 trials, achieving 73.5\% accuracy with spike-based input and output. These results validate that biologically inspired learning and quantization pipelines can translate effectively into low-power hardware implementations for cognitive state monitoring.

The implementation of the model using the analog circuits on the \ac{DYNAP-SE} was affected by the circuit device-mismatch, which induced variability in synaptic strength, membrane dynamics, and spike latencies. Despite this, the accuracy remained consistent across five trials ($\pm$0.6\%), indicating that appropriate network design and decoding strategies can mitigate hardware-induced fluctuations. Future work will investigate calibration methods and adaptive reconfiguration to enhance robustness over longer-term deployments.

\section{Conclusions}
\label{sec:conclusions}
This study presents a neuromorphic pipeline for cognitive state classification using spike-encoded multimodal physiological signals. We demonstrate that both conventional \ac{ML} models and \acp{SNN} achieve strong performance on a real-world \ac{ATC} task. Notably, a minimal single-layer \ac{SNN} achieved 80.6\% accuracy in software simulation and retained 73.5\% accuracy after quantization and deployment on the \ac{DYNAP-SE} neuromorphic chip.

To our knowledge, this is the first demonstration of real-time cognitive load classification from \ac{EEG} and eye-tracking features using quantized \acp{SNN} deployed on neuromorphic hardware. These findings validate the feasibility of minimal, biologically grounded models for embedded cognitive monitoring in dynamic real-world environments.

While conventional \ac{ML} approaches remain competitive baselines, \acp{SNN} offer unique advantages in energy efficiency and event-driven operation, making them well-suited for mobile and wearable applications.  Overall, this work highlights the potential of low-power, real-time neuromorphic systems for continuous mental state monitoring, with promising applications in aviation safety, workload tracking, and next-generation wearable neurotechnologies.

\section{Acknowledgements}
We acknowledge the financial support of the Digital Society Initiative (DSI) at University of Zurich.

\printbibliography
\vspace{12pt}
\color{red}

\end{document}